\documentclass[10pt,onecolumn,letterpaper]{article}

\usepackage{cvpr}
\usepackage{times}
\usepackage{epsfig}
\usepackage{graphicx}
\usepackage{amsmath}
\usepackage{amssymb}
\usepackage{algorithm}
\usepackage{amsthm}
\usepackage{algorithmic}
\usepackage{setspace}
\usepackage{subfigure}
\usepackage{graphicx}
\usepackage{multirow}
\usepackage{booktabs}
\usepackage{graphicx}
\usepackage[export]{adjustbox}

\usepackage[breaklinks=true,bookmarks=false]{hyperref}

\cvprfinalcopy 

\ifcvprfinal\fi

\begin{document}

\title{SoftMatch Distance: A Novel Distance for Weakly-Supervised Trend Change Detection in Bi-Temporal Images}

\author{Yuqun Yang, Xu Tang, Xiangrong Zhang, Jingjing Ma, Licheng Jiao}

\maketitle
\ifcvprfinal\thispagestyle{empty}\fi

\begin{abstract}
General change detection (GCD) and semantic change detection (SCD) are common methods for identifying changes and distinguishing object categories involved in those changes, respectively. However, the binary changes provided by GCD is often not practical enough, while annotating semantic labels for training SCD models is very expensive. Therefore, there is a novel solution that intuitively dividing changes into three trends (``appear'', ``disappear'' and ``transform'') instead of semantic categories, named it trend change detection (TCD) in this paper. It offers more detailed change information than GCD, while requiring less manual annotation cost than SCD. However, there are limited public data sets with specific trend labels to support TCD application. To address this issue, we propose a softmatch distance which is used to construct a weakly-supervised TCD branch in a simple GCD model, using GCD labels instead of TCD label for training. Furthermore, a strategic approach is presented to successfully explore and extract background information, which is crucial for the weakly-supervised TCD task. The experiment results on four public data sets are highly encouraging, which demonstrates the effectiveness of our proposed model.
	
\end{abstract}

\section{Introduction}
As a hot task in computer vision, change detection aims to identify the foreground (i.e., objects of interest) changes between bi-temporal images captured at the same locations~\cite{mandal2021empirical}. It has many real-world applications, such as land cover monitoring \cite{boguszewski2021landcover, mansour2020monitoring}, visual surveillance \cite{zaheer2021anomaly}, medical diagnosis \cite{haick2021artificial, shaheen2021adoption}, autonomous driving \cite{chen2021deep, zhou2021monocular}, etc. However, there are many difficulties need to be overcome. For example, affected by external conditions (e.g., illumination, sensor, and shooting angle), the visual appearance of the same/different foreground objects may be dissimilar/similar in bi-temporal images. At first, traditional machine learning techniques are popular for image change detection~\cite{vakalopoulou2015simultaneous, tm2015seamless, feng2015fine}. However, the low-level visual features limit their performance. Recently, assisted by the deep learning \cite{pang2021recorrupted, yang2022semiconductor}, change detection technologies are promoted significantly. Many deep-based methods \cite{park2022dual, hosseinzadeh2021image, verma2021qfabric} have been proposed and perform well in their applications.

\begin{figure}[t]
	\begin{center}
		\includegraphics[width=0.5\linewidth]{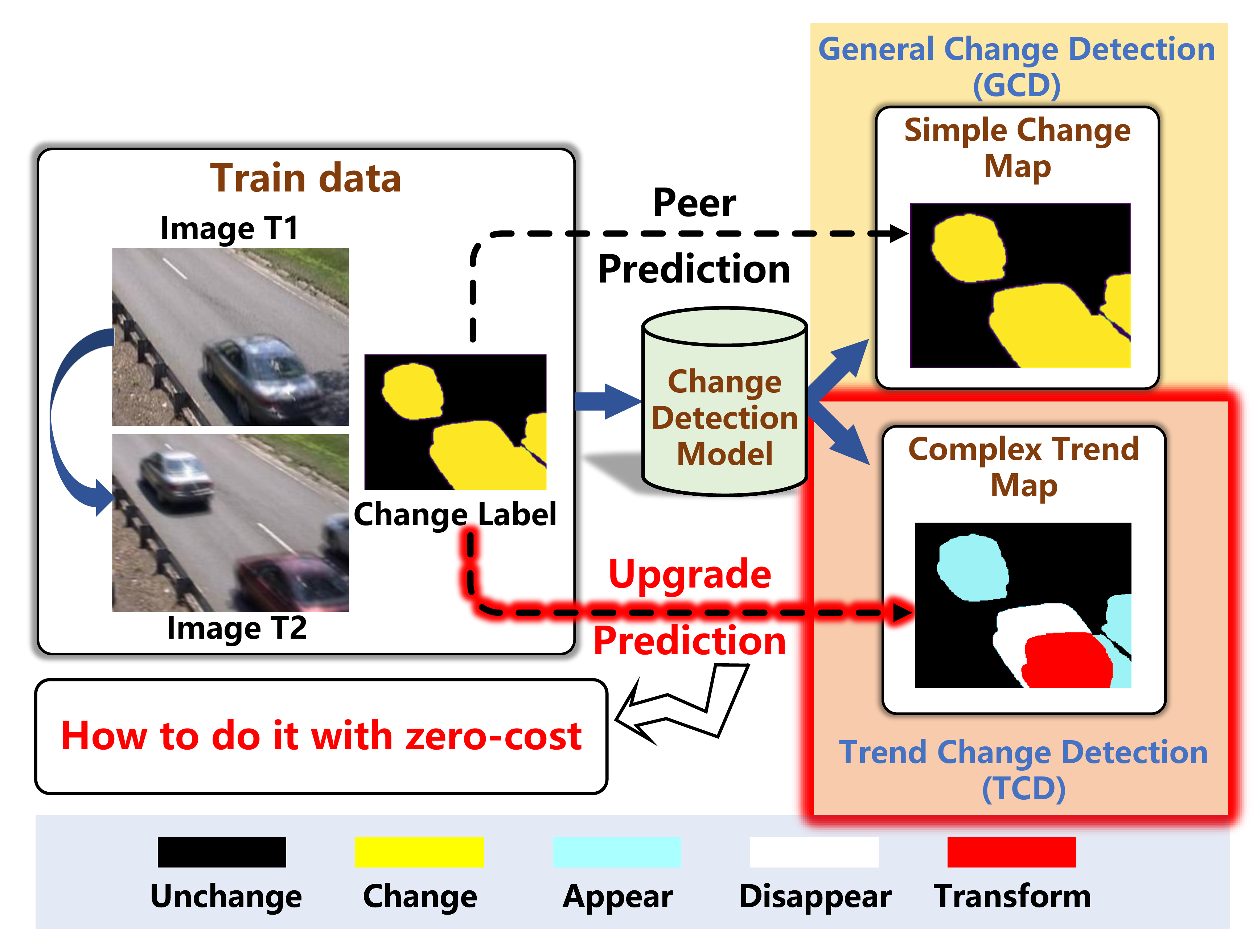}
	\end{center}
	\caption{Schematic illustration of main challenge. Based on the training of change label, most existing models conduct the peer prediction and output simple change map. However, how to realize the upgrade prediction for performing trend change detection with zero-cost for trend label annotation is a challenging problem, which also can be viewed as a weakly-supervised manner that using change labels instead of trend labels for training. We can find that the change map shows the changed car (marked in yellow). In contrast, the trend map further provides three trends: ``Appear'' (blue), ``Disappear'' (white) and ``Transform'' (red), offering a more detailed view of car changes.}
	\vspace{-5pt}
	\label{fig:task_defination}
\end{figure}

Generally, the existing deep-based models can be divided into two groups: general change detection (GCD) \cite{verma2021qfabric, vakalopoulou2015simultaneous, noh2022unsupervised} and semantic change detection (SCD) \cite{toker2022dynamicearthnet, zheng2022changemask, zheng2021change}. GCD models locate changes in foreground objects between bi-temporal images, while SCD models further recognize the semantic categories involved in those changes. However, both of them have respective limitations: 1) the outputs of GCD models, which only provide basic information on change or not, may be insufficient for practical use; and 2) SCD models require extensive manual labels for various semantics at pixel-level, which is very expensive. Although some unsupervised and weakly supervised SCD methods \cite{khoreva2017simple, caye2019guided, akiva2022self} have been proposed to alleviate this dilemma, the predicted results are always not satisfactory due to the absence of semantic labels. As a result, how to enhance the practicality of change detection and reduce the demands of manual labels has become an urgent and necessary problem. To address this issue, some methods \cite{park2021changesim, wang2023reduce} attempt to further divide changes into more intuitive and simple three trends, including ``Appear'', ``Disappear'' and ``Transform'' (see Fig. \ref{fig:task_defination}) instead of relying on complex semantic categories. These methods enrich the GCD results without massive semantic labels, and the obtained results could reflect the variation trend. Thus, we name them trend change detection (TCD) models in this paper. TCD is a relatively new concept that offers significant advantages for change detection in real-world applications, but lacks sufficient datasets with specific trend labels for support.

In this paper, we summarize a main challenge for this problem as illustrated in Fig. \ref{fig:task_defination}: how to realize the upgrade prediction to perform TCD task with zero-cost for annotating trend labels. Our solution is to propose a softmatch distance that is used to construct a weakly-supervised TCD branch in a simple GCD model. By utilizing only change labels of GCD, softmatch distance can equip features with the fore-/back-ground directivity (explained in Section \ref{section:softmatch}) for predicting trends. As a pluggable module, the softmatch distance has the potential to enable many existing GCD models for performing the weakly-supervised TCD task, significantly improving their practicalities.

Three main contributions can be summarized as follows:
\begin{itemize}
	\item To our knowledge, this is the first successful attempt in weakly-supervised TCD model, which is upgraded from a GCD model using softmatch distance. Moreover, we develop a strategic approach to accurately identify the important background information within bi-temporal images.
	\item We propose a softmatch distance that not only equips features with the fore-/back-ground directivity for the weakly-supervised TCD task, but also outperforms other common distances (i.e., cosine and Euclidean distances) in GCD task. 
	\item The extensive experiments are conducted on four public data sets to prove the effectiveness of the proposed method in performing GCD and weakly-supervised TCD tasks.
\end{itemize}

\section{Related Work}
\subsection{General Change Detection}
As a fundamental task, GCD plays a vital role in many applications \cite{mandal2021empirical}, and it always attracts researchers' attention. In \cite{subudhi2019kernelized}, a kernelized fuzzy modal variation method is proposed to use background subtraction to map video pixels into a high-dimensional space, allowing objects of interest to be separated from the backgrounds for obtaining change information. 
With the development of deep learning, many deep-based methods are developed based on the U-shape encoder-decoder networks \cite{wang2022uformer, noh2022unsupervised, osman2021transblast, alcantarilla2018street}. 
In \cite{lei2020hierarchical}, a hierarchical paired channel fusion network (HPCFNet) was introduced for street scene change detection. Incorporating channel fusion and multi-part feature learning between encoder and decoders improves its detection capacity. 
Another U-shape model for street scene change detection, named dynamic receptive temporal attention network (DRTAM), was proposed~\cite{chen2021dr}, which uses the temporal module to find an optimal dependency scope size for GCD dynamically. 
Furthermore, Park et al. \cite{park2022dual} developed SimSaC to enhance GCD model robustness in real-world settings. SimSaC carries out scene flow estimation and change detection simultaneously, detecting changes with imperfect matches.

Apart from the natural scenario, GCD is also a prevalent and significant task in the remote sensing (RS) community \cite{lv2021land, cheng2022isnet, li2022densely, fang2021snunet}. In literature \cite{daudt2018fully}, three U-shape models based on the fully convolutional network (FCN) were developed for RS GCD tasks, which adopt different feature fusion schemes to capture the multi-scale information hidden in RS images, guaranteeing the quality of GCD results. 
Tang et al. \cite{tang2021unsupervised} presented a GCD network for RS images using a Siamese FCN and graph convolutional network (GCN) to capture local information and long-distance pixel relationships, respectively. Then, change information is determined by a metric-learning algorithm.
Considering the high costs of labeling bi-temporal RS images, a single-temporal supervised learning method was proposed in \cite{zheng2021change}. It allows researchers to generate a high-performance change detection model only using unpaired label images.

\begin{figure*}[t]
	\begin{center}
		\includegraphics[width=1\linewidth]{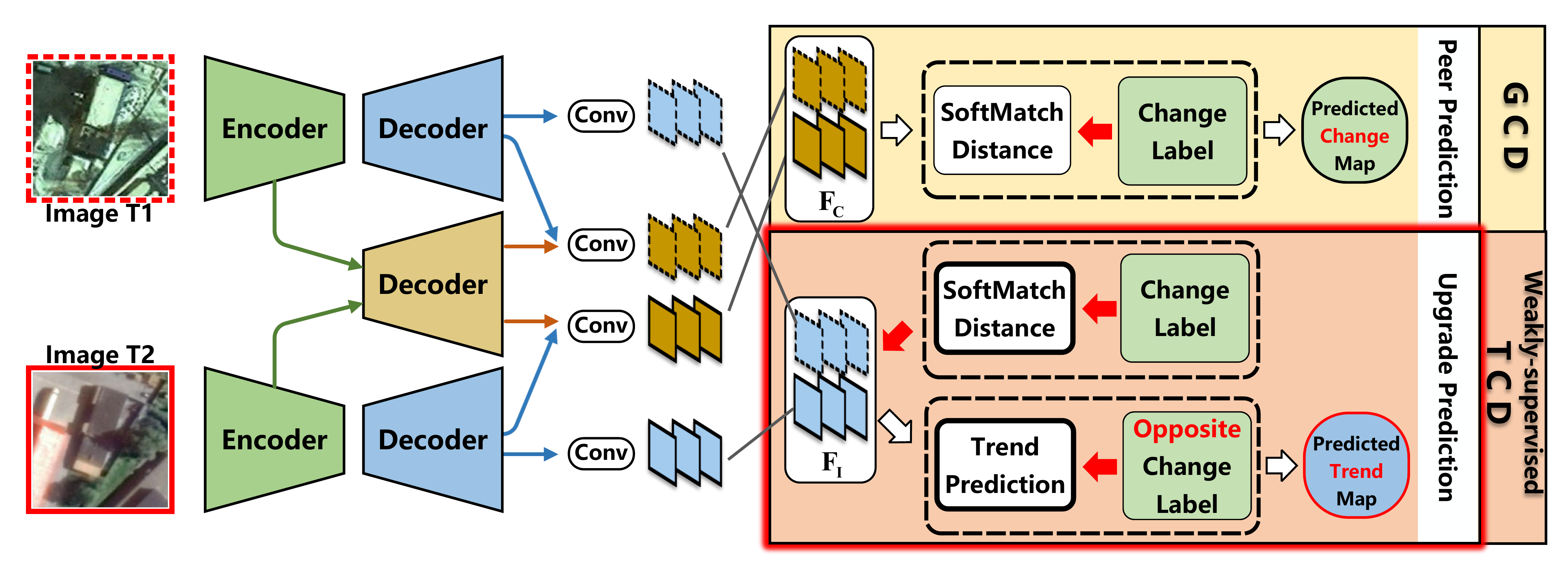}
	\end{center}
	\caption{Framework of the proposed method. An encoder-decoder is designed to extract two features groups: $\mathbf{F}_{\mathbf{C}}$ and $\mathbf{F}_{\mathbf{I}}$, which are applied to conduct the peer/upgrade prediction for GCD/TCD task with the output of change/trend map, respectively. In the GCD branch, the softmatch distance is used as distance function to measure the feature difference of $\mathbf{F}_{\mathbf{C}}$ for generating change map. In the weakly-supervised TCD branch, the softmatch distance can improve the fore-/back-ground directivity of $\mathbf{F}_{\mathbf{I}}$ under the supervision of change label. Meanwhile, the supervision of opposite change label is applied to explore the background information within bi-temporal images. Finally, the trend map is generated by the trend prediction module.}
	\label{fig:framework}
	\vspace{-5pt}
\end{figure*}
\subsection{Semantic Change Detection}
The key to SCD is recognizing the category information of the changed objects. Semantic segmentation models can intrinsically achieve this by comparing the semantic labels of pixels.\cite{caye2019guided}. For example,
inspired by human observation manner, Li et al.~\cite{li2022deep} proposed a hierarchical model for scene semantic segmentation that decomposes complex scenes into simpler parts and builds structured relations between them for obtaining accurate results.
Considering the importance of global semantic relations, a cross-image relational knowledge distillation strategy was developed in \cite{yang2022cross}, which transfers structured relations between pixel and region for favorable segmentation results. 

While semantic segmentation can be an emergency plan for SCD, it fails to consider the temporal relationships between multi-temporal images. Furthermore, identifying all pixels' semantic categories for SCD tasks is not always necessary. This is because the changed parts usually only occupy a fraction of the total image.
To overcome the above limitations, many SCD-oriented models have been developed based on the general-purpose semantic segmentation models. 
Ru et al. \cite{ru2020multi} presented a CorrFusion module for SCD. Based on regular convolutional neural network (CNN) and mapping techniques, CorrFusion can first get the bi-temporal images' representations and their instance-level correlation. Then, a cross-temporal fusion approach is used to explore the temporal relations and generate positive SCD results. 
A multi-task learning framework was introduced in \cite{ding2022bi} to deal with SCD. This framework consists of two temporal branches and a change branch. By enhancing their communications, the contributions of triple branches can be maximized. 
Peri Akiva et al. \cite{akiva2022self} proposed a material- and texture-based self-supervision method and applied it to SCD tasks. The reported results confirm its usefulness. The prerequisite of the above methods is that the labels are correct. 
However, noisy labels are inevitable in practice. To mitigate the negative influence caused by label noises, a new method was presented in \cite{caye2019guided}. On the one hand, an iterative learning model is developed to filter the noisy labels. On the other hand, a guided anisotropic diffusion algorithm is proposed to ensure the SCD results.

\section{Methodology}
The framework of the proposed method is shown in Fig. \ref{fig:framework}, which can be divided into three parts, i.e., feature learning, GCD branch and the weakly-supervised TCD branch. The feature learning part aims to fuse the valuable information extracted from the bi-temporal images to the paired common features $\mathbf{F}_\mathbf{C}$ and the paired independent features $\mathbf{F}_\mathbf{I}$. Under the supervision of change label, the GCD/TCD branch aim to generate change/trend maps using common/independent features $\mathbf{F}_\mathbf{C}$/$\mathbf{F}_\mathbf{I}$. Here, the cross-entropy function is used for all the supervisions.
\begin{figure}[t]
	\begin{center}
		\includegraphics[width=0.7\linewidth]{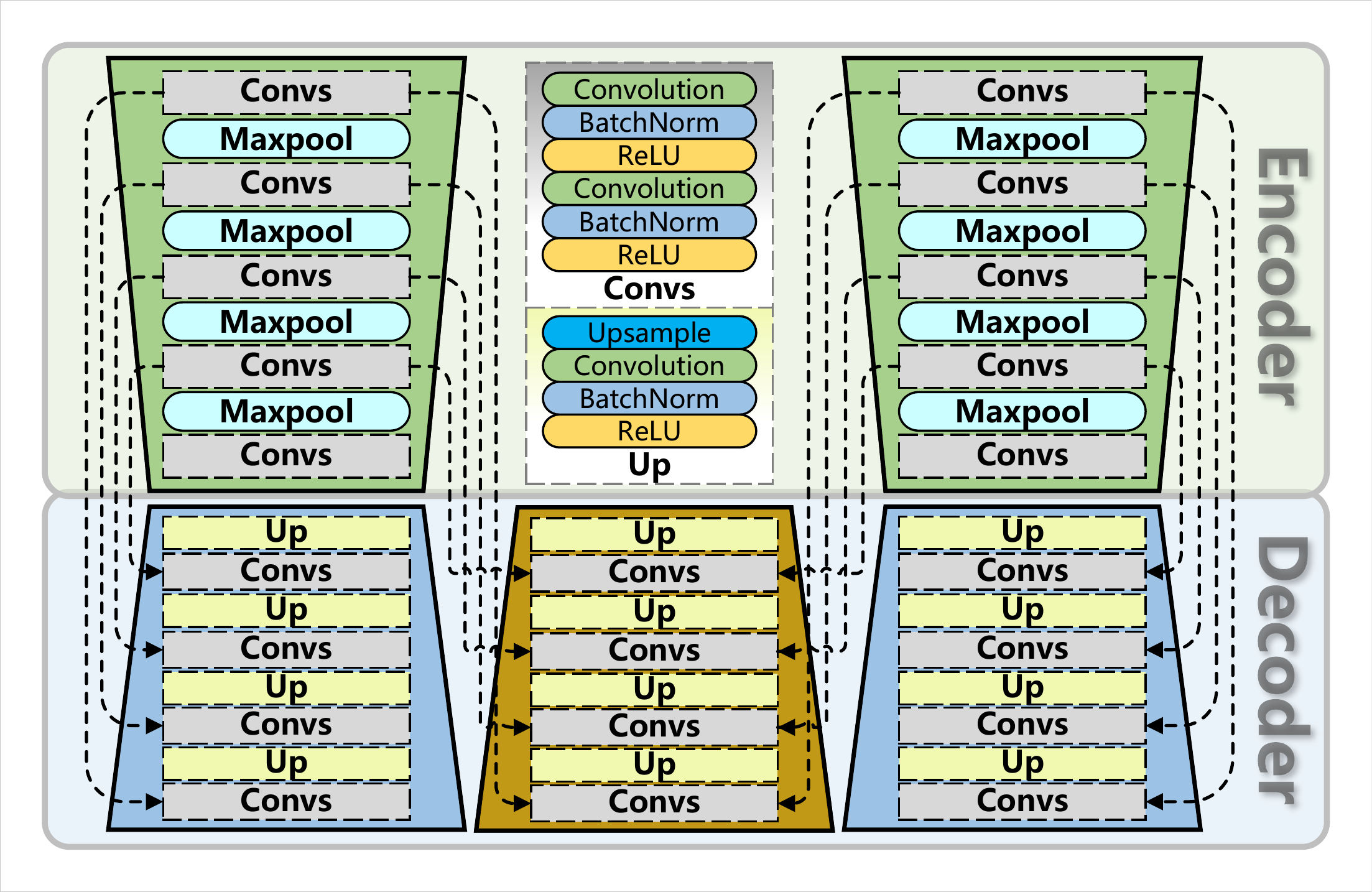}
	\end{center}
	\caption{Diagram of extracting features. It consists of two independent encoders (marked in green) and decoders (marked in blue) with shared parameters, and a common decoder (marked in brown). }
	\label{fig:feature_extraction}
	\vspace{-5pt}
\end{figure}

\subsection{Feature Extraction}
\label{feature_ex}
A U-shape encoder-decoder network is constructed to obtain the representative and discriminative visual features from the bi-temporal images (\textit{T1} and \textit{T2}), whose structure is exhibited in Fig. \ref{fig:feature_extraction}. Besides two same encoder-decoder nets (the left and right branches), which focus on capturing the profitable visual clues from input images, another decoder (the middle branch) is developed to extra explore temporal relationships between \textit{Image T1} and \textit{Image T2}.

Let us start with the explanation of two encoder-decoder nets. Since they have the same structure, we explain the left one (in Fig. \ref{fig:feature_extraction}) for clarity. The encoder contains five $Convs$ layers and four max-pooling layers with the two-times down-sampling. Note that a $Convs$ layer involves two groups of the convolution ($3\times 3$), BatchNorm, and ReLU. The decoder has four $Up$ layers and four $Convs$ layers. Here, a $Up$ layer incorporates an up-sampling, a convolution ($3\times 3$), a BatchNorm, and a ReLU. When \textit{Image T1} is input into the encoder, five multi-layer features $\mathbf{e}_{1}^{1}$, $\mathbf{e}_{2}^{1}$, $\mathbf{e}_{3}^{1}$, $\mathbf{e}_{4}^{1}$, and $\mathbf{e}_{5}^{1}$ can be produced. Then, $\mathbf{e}_{5}^{1}$ would be fed into the decoder to generate $\mathbf{d}_{5}^{1}$, $\mathbf{d}_{4}^{1}$, $\mathbf{d}_{3}^{1}$, and $\mathbf{d}_{2}^{1}$. Here, common skip connections are added between the encoder and decoder to integrate the various knowledge within the multi-layer features and reduce information loss. Thus, the features of the decoder are output by,
\begin{equation}
	\label{feature_fusion}
	\begin{aligned}
		\mathbf{d}_{i}^{1}=\left\{ \begin{matrix}
			\operatorname{Convs}\left( \operatorname{Up}\left( \mathbf{e}_{i}^{1} \right)\oplus \mathbf{e}_{i-1}^{1} \right),i=5  \\
			\operatorname{Convs}\left( \operatorname{Up}\left( \mathbf{d}_{i+1}^{1} \right)\oplus \mathbf{e}_{i-1}^{1} \right),i=4,3,2  \\
		\end{matrix} \right.,
	\end{aligned}
\end{equation}
where $\operatorname{Convs}\left( \cdot  \right)$, $\operatorname{Up}\left( \cdot  \right)$, and $\oplus$ indicate the nonlinear functions of the $Convs$ layer and $Up$ layer, and concatenation, respectively. Note that the spatial size of $\mathbf{d}_{2}^{1}$ is the same as that of \textit{Image T1}. Similarly, after passing through the encoder-decoder net, \textit{Image T2} can be transformed into $\left\{ \mathbf{e}_{1}^{2},\mathbf{e}_{2}^{2},\mathbf{e}_{3}^{2},\mathbf{e}_{4}^{2},\mathbf{e}_{5}^{2} \right\}$ and $\left\{ \mathbf{d}_{5}^{2},\mathbf{d}_{4}^{2},\mathbf{d}_{3}^{2},\mathbf{d}_{2}^{2} \right\}$, where the spatial sizes of ${{\bf{d}}_2^2}$ and \textit{Image T2} are the same to each other. Based on the obtained features, we can get the paired independent features ${{\mathbf{F}}_{\mathbf{I}}}\text{=}\left\{ \mathbf{F}_{\mathbf{I}}^{1},\mathbf{F}_{\mathbf{I}}^{2} \right\}$ by,
\begin{equation}
	\begin{aligned}
		{{\mathbf{F}}_{\mathbf{I}}}=\left\{ \operatorname{conv}_{1\times1}\left( \mathbf{d}_{2}^{1} \right),\operatorname{conv}_{1\times1}\left( \mathbf{d}_{2}^{2} \right) \right\},
	\end{aligned}
\end{equation}
where the output channel of $\operatorname{conv}_{1\times1}\left( \cdot  \right)$ is $c$.

As mentioned above, there is another decoder (see Fig. \ref{fig:feature_extraction}, the middle one) for modeling the temporal relationship. To this end, the multi-layer features $\mathbf{e}_{i}^{1}$ and $\mathbf{e}_{i}^{2}$ of \textit{Image T1} and \textit{Image T2} will be fused in this decoder. Firstly, $\mathbf{e}_{i}^{1}$ and $\mathbf{e}_{i}^{2}$ are concatenated in channel dimension to generate $\mathbf{e}_{i}^{c}$. Then, the same fusion scheme with Equation \ref{feature_fusion} is adopted to fuse $\mathbf{e}_{i}^{c}$ and output $\mathbf{d}_{2}^{c}$.
Finally, we can get the paired common features ${{\mathbf{F}}_{\mathbf{C}}}=\left\{ \mathbf{F}_{\mathbf{C}}^{1},\mathbf{F}_{\mathbf{C}}^{2} \right\}$ by,
\begin{equation}
	\begin{aligned}
		{{\mathbf{F}}_{\mathbf{C}}} = \left\{ {{\mathop{\rm conv}_{1\times1}\nolimits} \left( {{\bf{d}}_2^1 \oplus {\bf{d}}_2^c} \right),{\mathop{\rm conv}_{1\times1}\nolimits} \left( {{\bf{d}}_2^2 \oplus {\bf{d}}_2^c} \right)} \right\},
	\end{aligned}
\end{equation}
where the output channel of $\operatorname{conv}_{1\times1}\left( \cdot  \right)$ also equals $c$. Note that we set $c=3$, and Section \ref{section:Trend_prediction} will explain the reasons.

\subsection{SoftMatch Distance}
\label{section:softmatch}
Usually, after obtaining the paired bi-temporal features $\left\{ \mathbf{F}_{\centerdot}^{1},\mathbf{F}_{\centerdot}^{2} \right\}$, we can adopt some common distance metrics (e.g., Cosine distance \cite{wang2018cosface}) to measure their difference for generating change map, which is effective in our GCD branch. However, due to the specific goals of our weakly-supervised TCD branch, in addition to the capacity of measuring feature difference, we also require the distance metric can equip fore-/back-ground directivity to the involved features based on the supervision of change label. In other words, we expect that the final $\left\{ \mathbf{F}_{\centerdot}^{1},\mathbf{F}_{\centerdot}^{2} \right\}$ (after training) can provide specific information to distinguish different foregrounds or the foreground and background between bi-temporal images, in order to predict trend map. In addition, we hope the values of changed/unchanged pixels in the generated distance maps (DMs) can be pushed toward 1/0 for aligning with the change label when computing loss. 


\begin{figure}[h]
	\begin{center}
		\includegraphics[width=0.7\linewidth]{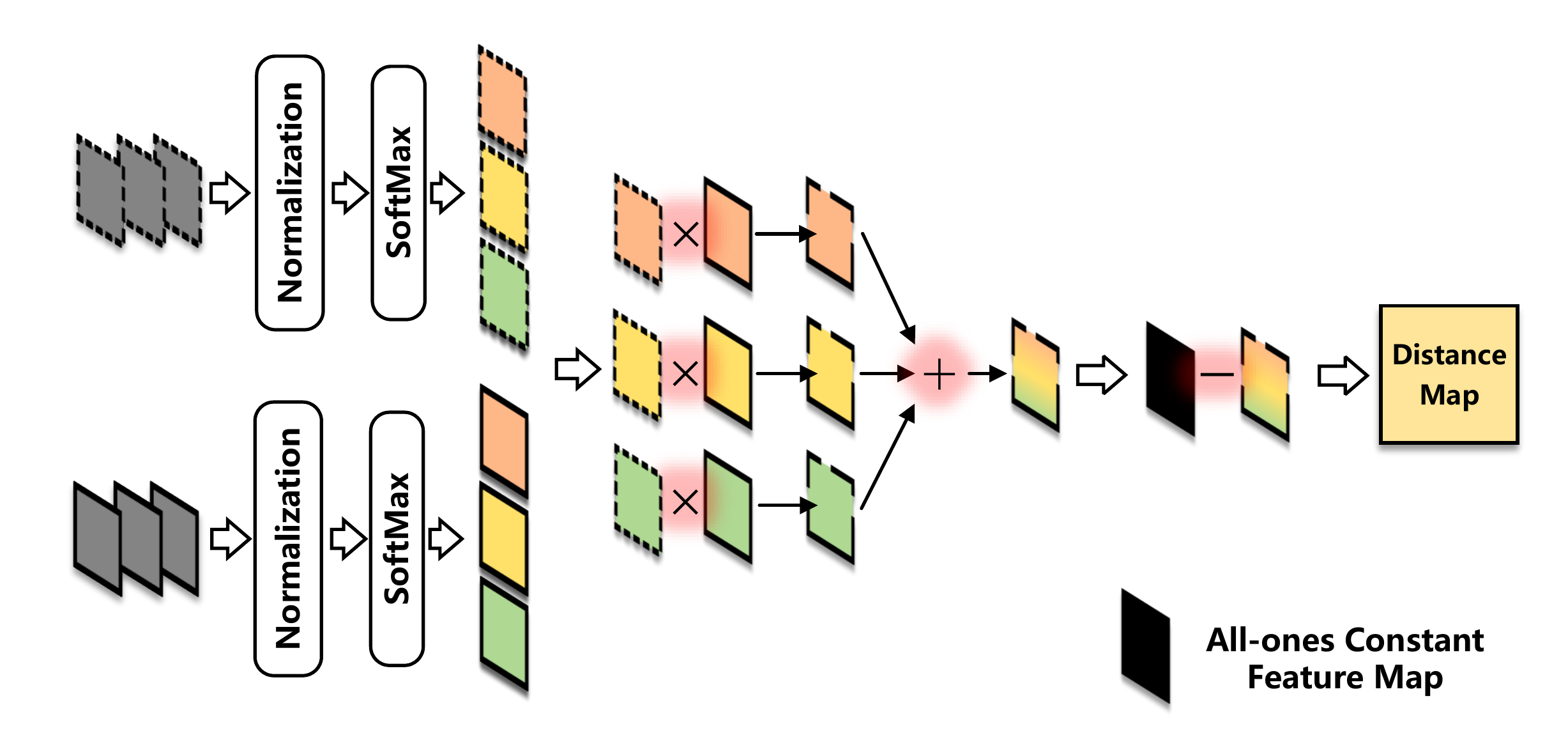}
	\end{center}
	\caption{Diagram of computing softmatch distance between two feature maps. Here, all of the ``$\times$'', ``$+$'', and ``$-$'' operations are carried out at the pixel-level.}
	\label{fig:softmatch}
	\vspace{-5pt}
\end{figure}

Summing up the above factors that are not achievable with existing common distances, we develop a simple yet effective softmatch distance. Here,
${{\mathbf{F}}_{\mathbf{I}}} = \left\{ \mathbf{F}_{\mathbf{I}}^{1},\mathbf{F}_{\mathbf{I}}^{2} \right\}$ is used as an example to illustrate the proposed softmatch distance, while the computation for ${{\mathbf{F}}_{\mathbf{C}}} = \left\{ \mathbf{F}_{\mathbf{C}}^{1},\mathbf{F}_{\mathbf{C}}^{2} \right\}$ is the same. 
To calculate the softmatch distance of feature pixels $\mathbf{p}=\{\mathbf{p}^1, \mathbf{p}^2\}, \mathbf{p}^i \in \mathbb{R}^{c}$ within ${{\mathbf{F}}_{\mathbf{I}}}$, the normalization of subtracting the maximum value and the softmax operation are applied first to obtain $\mathbf{s}=\left\{ {{\mathbf{s}}^{1}},{{\mathbf{s}}^{2}} \right\},{{\mathbf{s}}^{i}}\in {{\mathbb{R}}^{c}}$. Then, subtracting $1$ from the inner product of ${{\mathbf{s}}^{1}}$ and ${{\mathbf{s}}^{2}}$ is conducted, i.e.,
\begin{equation}
	\begin{aligned}
		\begin{matrix}
			sdist\left( {{\mathbf{p}}^{1}},{{\mathbf{p}}^{2}} \right)=1-\left\langle \mathbf{s}_{i}^{1},\mathbf{s}_{i}^{2} \right\rangle   \\
			=1-\sum\limits_{i=0}^{c}{\left( \frac{\exp \left( p_{i}^{1} \right)}{\sum\limits_{j=1}^{c}{\exp \left( p_{j}^{1} \right)}}\times \frac{\exp \left( p_{i}^{2} \right)}{\sum\limits_{j=1}^{c}{\exp \left( p_{j}^{2} \right)}} \right)}.  \\
		\end{matrix}
	\end{aligned}
\end{equation}
Here, the \textit{Proof} that the value range of $sdist\left( {{\mathbf{p}}^{1}},{{\mathbf{p}}^{2}} \right)$ is $(0,1)$ is provided in supplementary materials for clearly measuring feature differences. When all pixels' softmatch distances are calculated, the DM of $\left\{ \mathbf{F}_{\mathbf{I}}^{1},\mathbf{F}_{\mathbf{I}}^{2} \right\}$ can be obtained. In the same way, the DM of $\left\{ \mathbf{F}_{\mathbf{C}}^{1},\mathbf{F}_{\mathbf{C}}^{2} \right\}$ can be calculated. The diagram of calculating the softmatch distance between two features is shown in Fig. \ref{fig:softmatch}.

After obtaining DMs, the model parameters can be optimized to perform their respective tasks by computing the loss between DMs and change labels. Then, in the GCD branch, it is intuitive that the DM output by the trained model is discretized to be the predicted change map. Now, let's delve into how the fore-/back-ground directivity of $\mathbf{F}_{\mathbf{I}} = \left\{ \mathbf{F}_{\mathbf{I}}^{1},\mathbf{F}_{\mathbf{I}}^{2} \right\}$ is obtained in the weakly-supervised TCD branch. 

\begin{figure}[h]
	\begin{center}
		\includegraphics[width=0.7\linewidth]{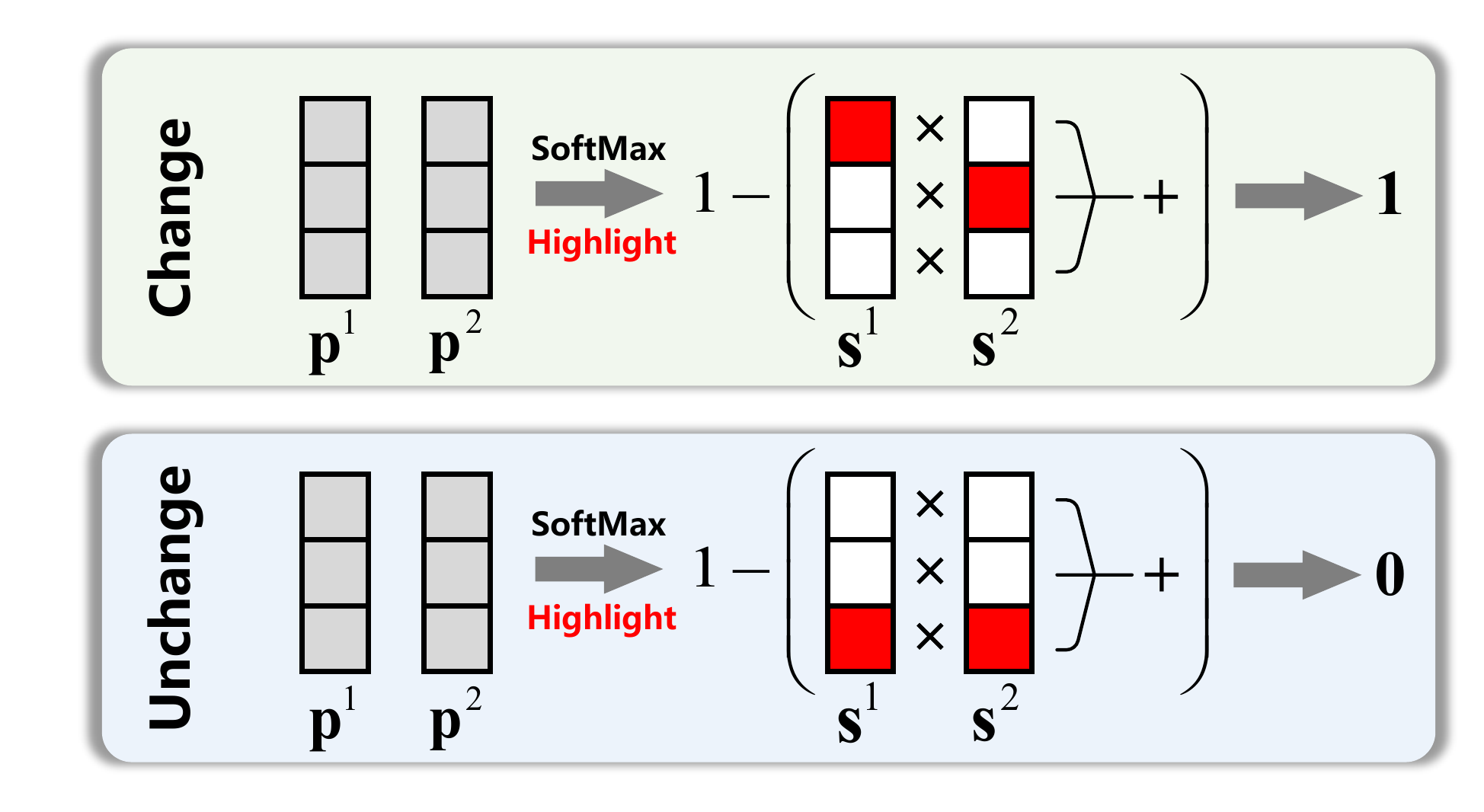}
	\end{center}
	\caption{Example of highlighting one channel of $\mathbf{s}^1$ and $\mathbf{s}^2$. Here, the values of highlighted/unhighlighted channel (mark in red/white) are close to 1/0 due to using softmax operation.}
	\label{fig:highlight}
	\vspace{-7pt}
\end{figure}

For clarity, we will use a paired pixels $\mathbf{p}=\{\mathbf{p}^1, \mathbf{p}^2\}$ within $\mathbf{F}_{\mathbf{I}}$ as an example to illustrate this point. In fact, the fore-/back-ground directivity is reflected in that each channel of $\mathbf{p}^1$ and $\mathbf{p}^2$ is associated with a particular foreground or background. In this way, once a pixel channel is highlighted, it becomes clear which foreground or background it belongs to, enabling the trend map generation. To do this, we first leverage $\mathbf{s}=\{\mathbf{s}^1, \mathbf{s}^2\}$ output by softmax operation instead of $\mathbf{p}=\{\mathbf{p}^1, \mathbf{p}^2\}$ to ensure two points: 1) at most one channel can be highlighted; and 2) the highlighted/unhighlighted values are constrained to be no greater/less than 1/0. Although we have defined two limitations for normalizing the feature values, it is still unclear how to push the highlighted/unhighlighted values toward 1/0 for certainly directivity. Fortunately, applying two operations, inner product and subtraction of $1$, to $\mathbf{s}=\{\mathbf{s}^1, \mathbf{s}^2\}$ can achieve this goal. As shown in Fig. \ref{fig:highlight}, for the scenarios of changed/unchanged, model has to highlight different/same channels of $\mathbf{s}^1$ and $\mathbf{s}^2$ and push their values toward 1 for generating the result value of 1/0. 

After training, an anonymous association plan between feature channels and foreground/background is created by $\mathbf{s}^1$ and $\mathbf{s}^2$, with one channel highlighted to indicate its membership. Therefore, the fore-/back-ground directivity of $\mathbf{F}_{\mathbf{I}}$ is obtained successfully during training. However, the current functions are insufficient to preform TCD task due to a key problem: the anonymous association plan does not clearly indicate which channel corresponds to the foreground or the background. Its solution to expose this plan will be provided in Section \ref{section:Trend_prediction}.

Finally, let's look at three possible concerns due to applying similar manner to conduct softmatch distance in both TCD and GCD branches. \textbf{Q1}) Is the fore-/back-ground directivity of ${{\mathbf{F}}_{\mathbf{C}}}$ also obtained in the GCD branch?  \textbf{A1}) Yes, but the common information within $\mathbf{F}_{\mathbf{C}}$ interferes with its ability to distinguish foreground and background for predicting trend. Nonetheless, it helps softmatch distance achieve more accurate measurements than common distances, thereby improving performance of GCD. \textbf{Q2}) Why isn't the DM of $\mathbf{F}_{\mathbf{I}}$ in the TCD branch used to generate change maps? \textbf{A2}) This is unnecessary since the predicted trend map already includes both change and trend information. \textbf{Q3}) Why not directly use the trend map of TCD branch to generate the change map, instead of saving the GCD branch? \textbf{A3}) The trend map can reflect changes, but since its supporting features $\mathbf{F}_{\mathbf{I}}$ lack common temporal information, the change map generated by it may be slightly inferior to that generated by $\mathbf{F}_{\mathbf{C}}$.

\subsection{Trend Prediction}
\label{section:Trend_prediction}

As mentioned earlier, softmatch distance alone only establish the unclear anonymous association plan, which is not enough for weakly-supervised TCD. Therefore, to expose it, we need to further analyze how to definitely assign $c$ channels of  $\mathbf{F}_{\mathbf{I}}$ to foregrounds and the background, in order to meet the trend prediction requirements.


Again, we define three trends ``appear'', ``disappear'', and ``transform'', i.e.,
\begin{itemize}
	\item ``Appear'': from background to foreground,
	\item ``Disappear'': from foreground to background,
	\item ``Transform'': from foreground A to foreground B.
\end{itemize}
Where the foreground tags A and B are not specific object categories, but instead are used to differentiate between two different foregrounds. By analyzing of above definitions, we can find that identifying background is paramount in TCD. Thus, only one of $c$ channels of  $\mathbf{F}_{\mathbf{I}}^{1}$ will be selected to learn the background, and we name it the background channel. The rest $c-1$ channels will be regarded as the foreground channels, which focus on discovering the objects of interest. Here, $c-1$ is set to the minimal value $2$ of satisfying ``Transformation'', i.e., $c=3$. The same actions would be carried out to $\mathbf{F}_{\mathbf{I}}^{2}$. Thus, the three channels of $\mathbf{F}_{\mathbf{I}}^{1}$ and $\mathbf{F}_{\mathbf{I}}^{2}$ correspond to foreground A, foreground B, and background. Here, only the background channel needs to be associated fixedly with one channel, while the foreground channels can be associated adaptively as they only indicate the difference between foregrounds. Therefore, fixing the background channel allows us to confirm the exact channel association plan for performing TCD task.

To fix background channel of $\mathbf{F}_{\mathbf{I}}$, because most background areas would be unchanged, an opposite change label (change/unchanged pixels are annotated as 0/1) is constructed to train one self-determined channel of $\mathbf{F}_{\mathbf{I}}$ using the cross-entropy loss function. Particularly, a strategic approach for accurately identifying background needs to be emphasized. The integrated cross-entropy loss will push the probability value of positive (unchanged) and negative (changed) samples toward 1 and 0 for representing background and non-background, respectively. However, this is not appropriate for negative samples, since the background frequently occurs in the changed parts of ``Appear'' and ``Disappear''. Thus, we only count the positive samples' loss for training model. Here, no action is required for the foreground channels.

\begin{figure}[h]
	\begin{center}
		\includegraphics[width=1\linewidth]{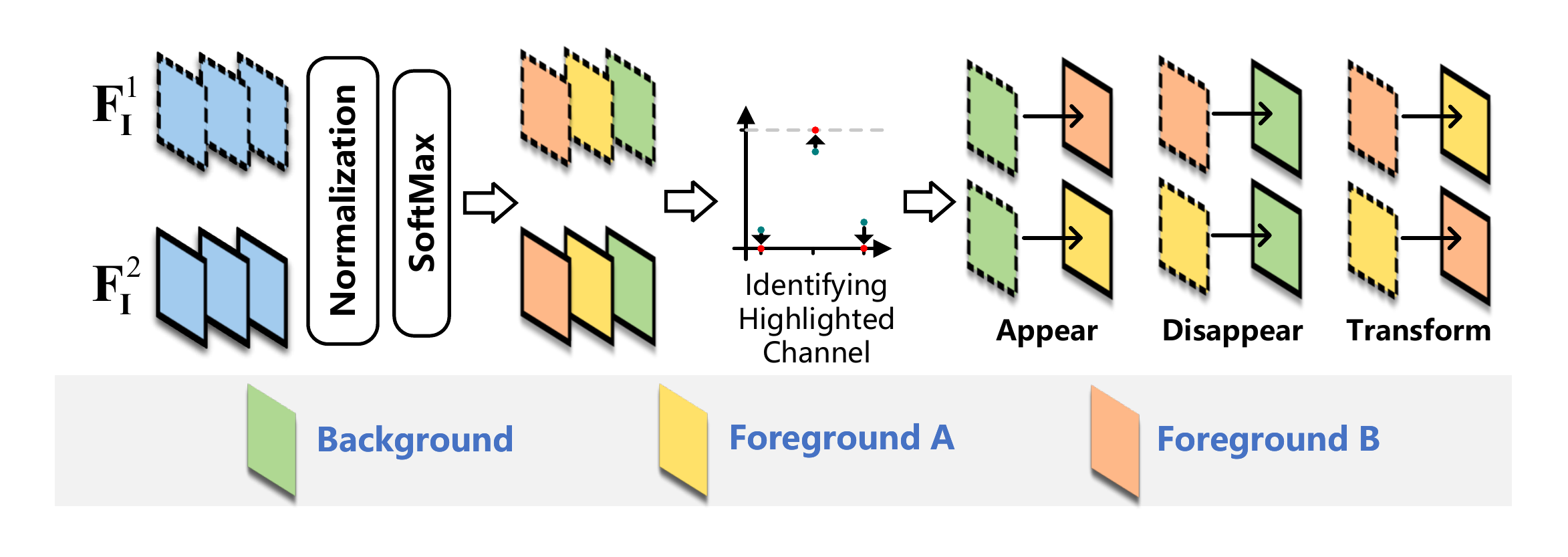}
	\end{center}
	\vspace{-5pt}
	\caption{Diagram of trend prediction in the inference stage.}
	\label{fig:trend_prediction}
\end{figure}

As shown in Fig. \ref{fig:trend_prediction}, when $\mathbf{F}_{\mathbf{I}}^{1}$ and $\mathbf{F}_{\mathbf{I}}^{2}$ are trained, the trend change maps can be generated by combining the following three cases. ``Appear''/``Disappear'' can be represented by the highlighted background/foreground channel of $\mathbf{F}^1_\mathbf{I}$ is converted to the highlighted foreground/background channel of $\mathbf{F}^2_\mathbf{I}$. ``Transform'' is denoted as that the highlighted foreground A/B channel of $\mathbf{F}^1_\mathbf{I}$ is converted to the highlighted foreground B/A channel of $\mathbf{F}^2_\mathbf{I}$.

\begin{figure*}[t]
	\centering
	\hspace{0cm}
	\subfigure[Image \textit{T1}]{
		\begin{minipage}[b]{0.10\linewidth}
			\centering
			\includegraphics[width=0.65in]{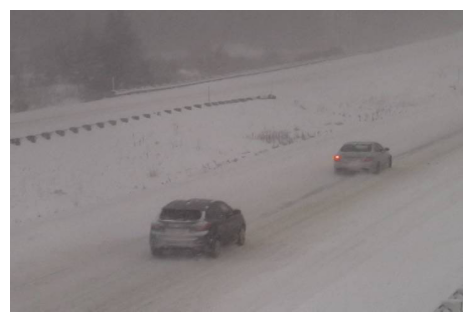}\hspace{5pt}
			\includegraphics[width=0.65in]{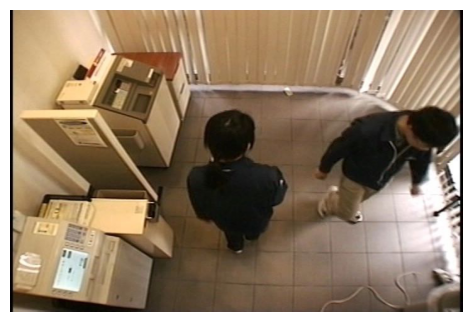}\hspace{5pt}
			\includegraphics[width=0.65in]{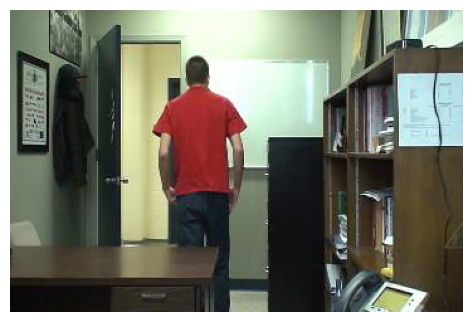}\hspace{5pt}
			\includegraphics[width=0.65in]{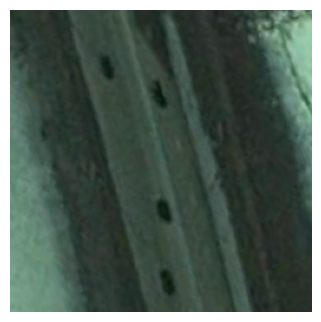}\hspace{5pt}
			\includegraphics[width=0.65in]{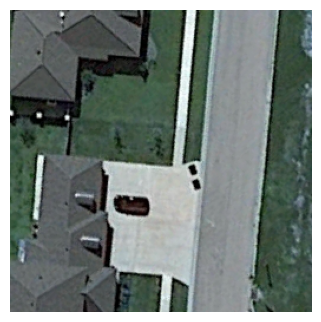}
			\includegraphics[width=0.65in]{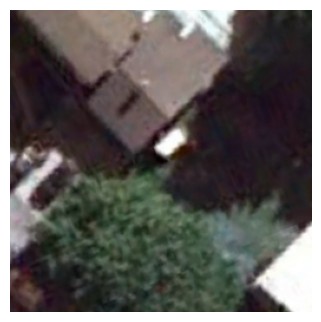}
			
		\end{minipage}
	}
	\subfigure[Image \textit{T2}]{
		\begin{minipage}[b]{0.10\linewidth}
			\centering
			\includegraphics[width=0.65in]{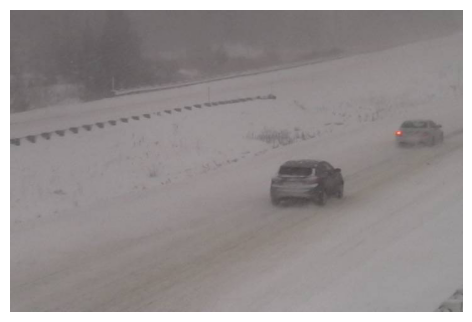}\hspace{5pt}
			\includegraphics[width=0.65in]{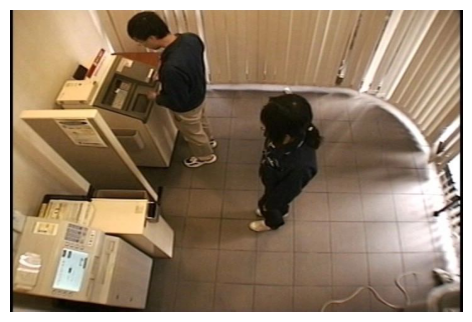}\hspace{5pt}
			\includegraphics[width=0.65in]{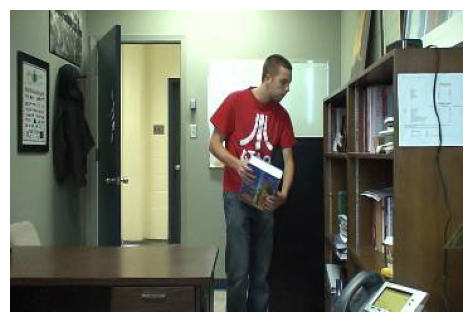}\hspace{5pt}
			\includegraphics[width=0.65in]{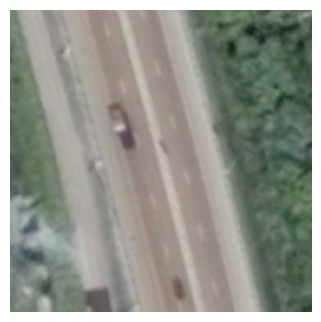}\hspace{5pt}
			\includegraphics[width=0.65in]{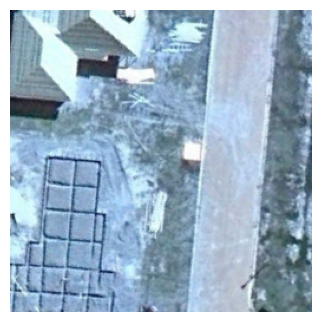}
			\includegraphics[width=0.65in]{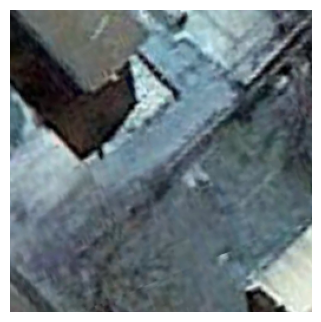}
			
		\end{minipage}
	}
	\subfigure[{Change label}]{
		\begin{minipage}[b]{0.10\linewidth}
			\centering
			\includegraphics[width=0.65in]{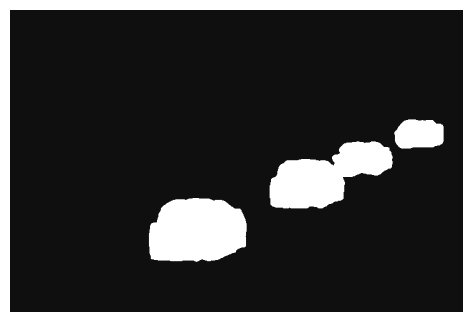}\hspace{5pt}
			\includegraphics[width=0.65in]{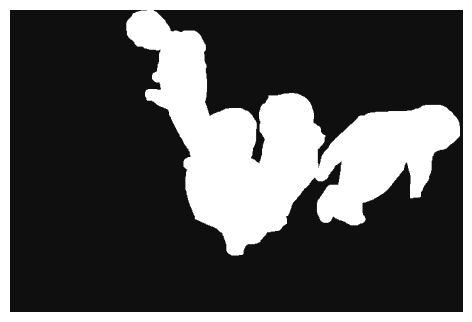}\hspace{5pt}
			\includegraphics[width=0.65in]{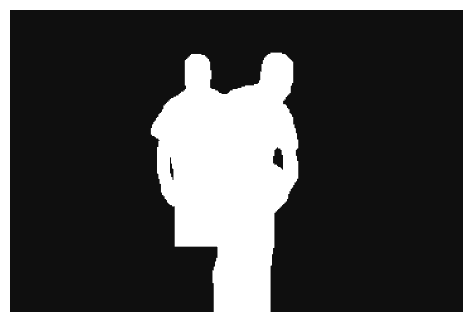}\hspace{5pt}
			\includegraphics[width=0.65in]{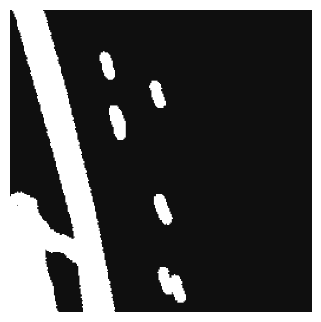}\hspace{5pt}
			\includegraphics[width=0.65in]{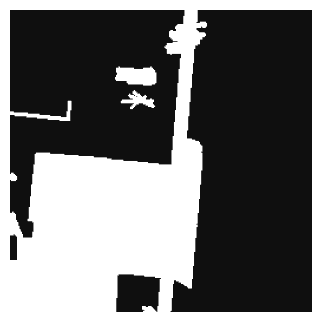}
			\includegraphics[width=0.65in]{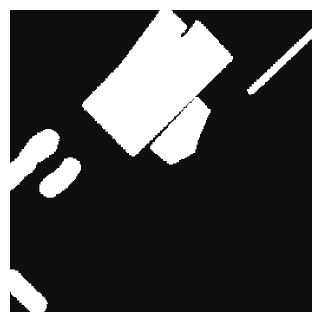}
		\end{minipage}
	}
	\subfigure[Trend label]{
	\begin{minipage}[b]{0.10\linewidth}
		\centering
		\includegraphics[width=0.65in]{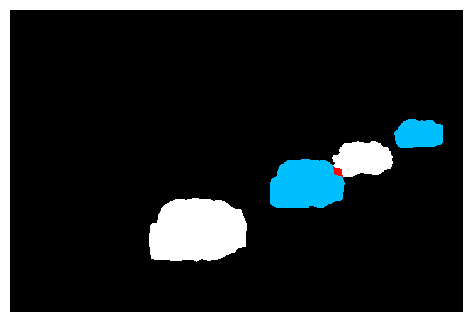}\hspace{5pt}
		\includegraphics[width=0.65in]{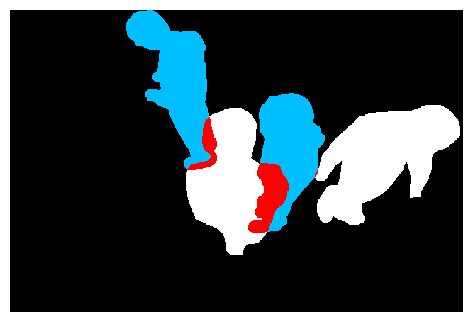}\hspace{5pt}
		\includegraphics[width=0.65in]{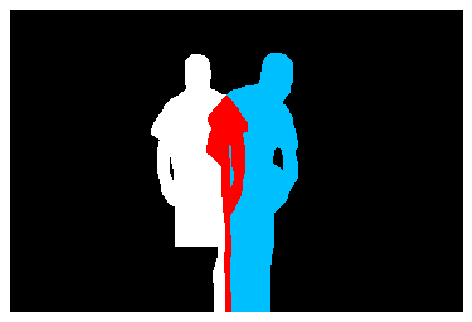}\hspace{5pt}
		\vspace{132.5pt}
	\end{minipage}
	}\ \ \ \ \ \ 
	\subfigure[Change map]{
		\begin{minipage}[b]{0.10\linewidth}
			\centering
			\includegraphics[width=0.65in]{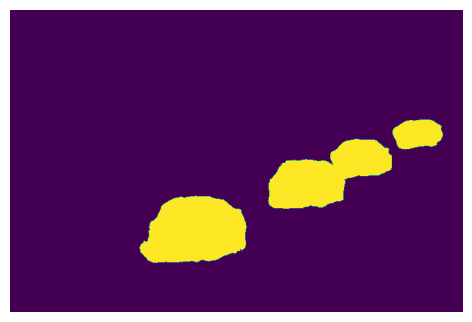}\hspace{5pt}
			\includegraphics[width=0.65in]{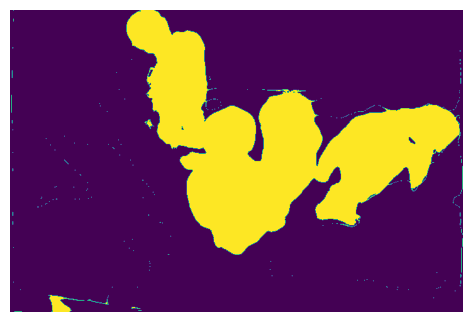}\hspace{5pt}
			\includegraphics[width=0.65in]{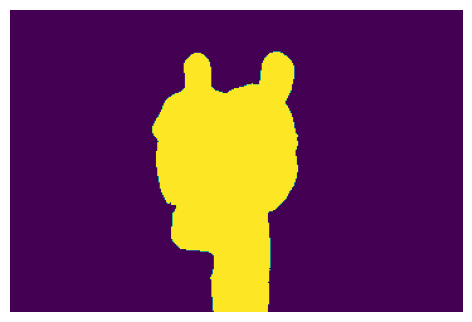}\hspace{5pt}
			\includegraphics[width=0.65in]{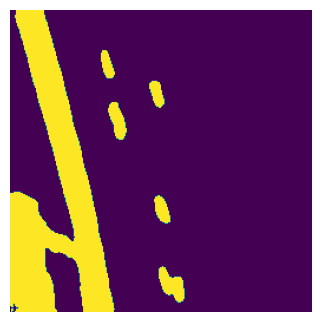}\hspace{5pt}
			\includegraphics[width=0.65in]{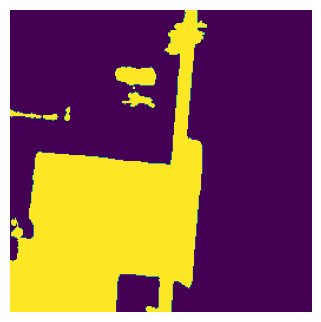}
			\includegraphics[width=0.65in]{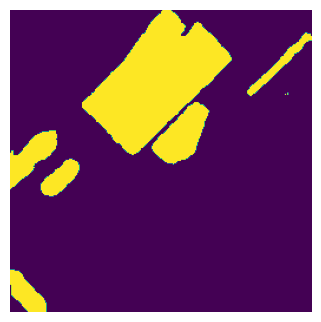}
			
		\end{minipage}
	}
	\subfigure[Trend map]{
		\begin{minipage}[b]{0.10\linewidth}
			\centering
			\includegraphics[width=0.65in]{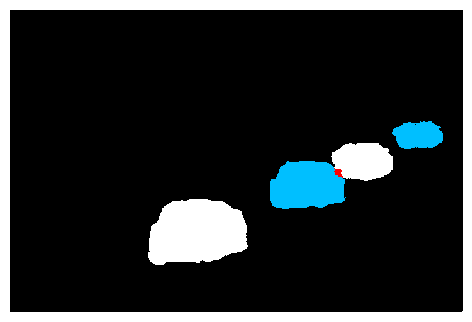}\hspace{5pt}
			\includegraphics[width=0.65in]{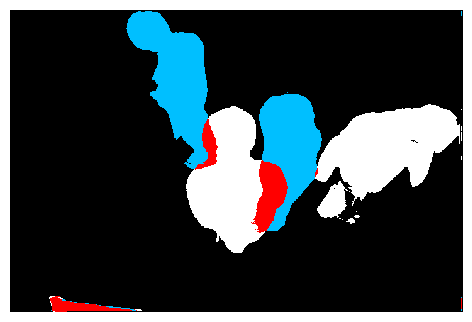}\hspace{5pt}
			\includegraphics[width=0.65in]{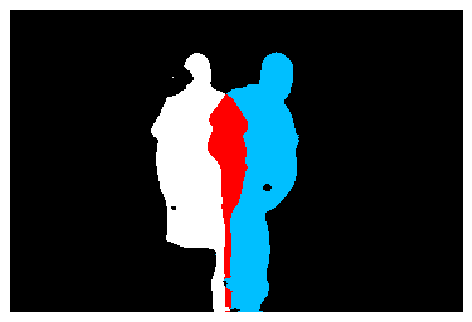}\hspace{5pt}
			\includegraphics[width=0.65in]{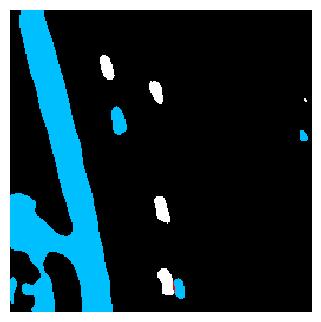}\hspace{5pt}
			\includegraphics[width=0.65in]{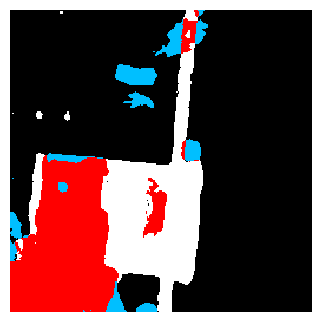}
			\includegraphics[width=0.65in]{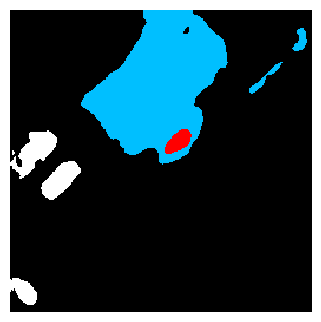}
		\end{minipage}
	}\ \ \ \ \ \ 
	\subfigure[\textit{T1} background]{
		\begin{minipage}[b]{0.115\linewidth}
			\label{feature_show_t1}
			\centering
			\includegraphics[width=0.65in]{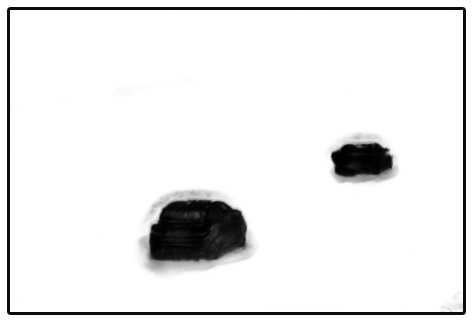}\hspace{5pt}
			\includegraphics[width=0.65in]{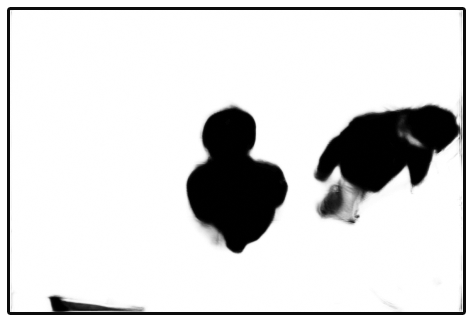}\hspace{5pt}
			\includegraphics[width=0.65in]{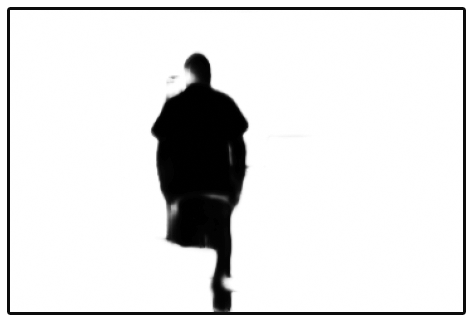}\hspace{5pt}
			\includegraphics[width=0.65in]{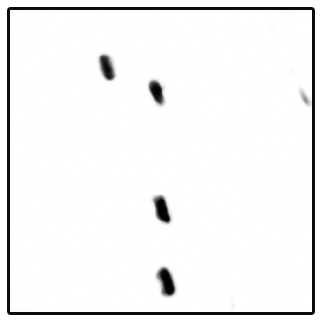}\hspace{5pt}
			\includegraphics[width=0.65in]{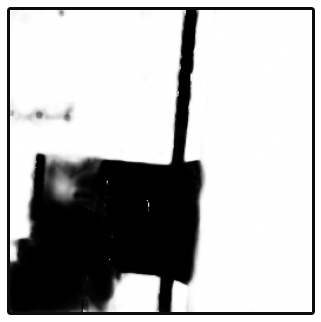}
			\includegraphics[width=0.65in]{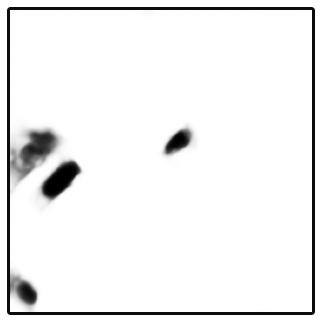}
			
		\end{minipage}
	}
	\subfigure[\textit{T2} background]{
		\begin{minipage}[b]{0.115\linewidth}
			\label{feature_show_t2}
			\centering
			\includegraphics[width=0.65in]{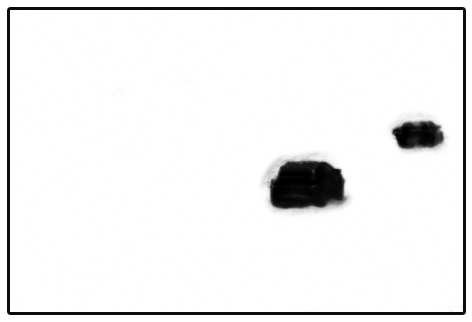}\hspace{5pt}
			\includegraphics[width=0.65in]{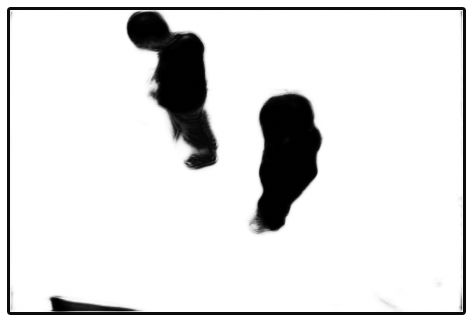}\hspace{5pt}
			\includegraphics[width=0.65in]{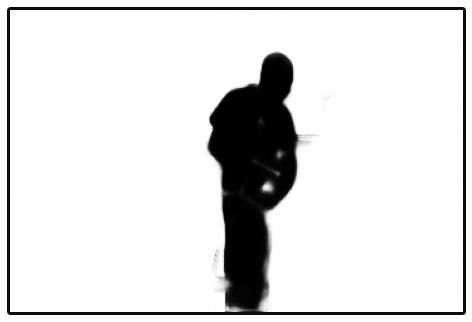}\hspace{5pt}
			\includegraphics[width=0.65in]{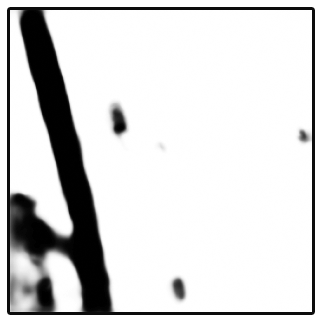}\hspace{5pt}
			\includegraphics[width=0.65in]{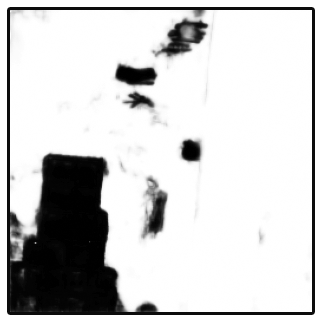}
			\includegraphics[width=0.65in]{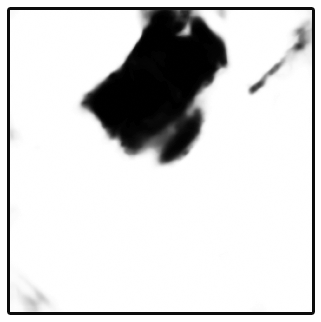}
		\end{minipage}
	}

	\caption{Six examples of conducting weakly-supervised TCD task, in which the first three lines and the last three lines are from CDnet-2014 and SVCD, respectively. (a), (b), (c) and (d) show the inputted \textit{image} $\textit{T}_1$, \textit{image} $\textit{T}_2$, change and trend labels. The predicted change and trend maps are exhibited in (e) and (f). As the SVCD dataset is presented for GCD task, it does not provide trend labels. It is worth noting that the background channel of $\mathbf{F}^1_\mathbf{I}$ and $\mathbf{F}^2_\mathbf{I}$ are shown in (g) and (h) to visualize extracted background information (marked in white).}
	\label{fig:feature_show}
	\vspace{-5pt}
\end{figure*}

\section{Experiments}
\subsection{Data Sets and Metrics}
To evaluate the model performance, four large-scale public change detection data sets are employed, including a natural image data set CDnet-2014 \cite{goyette2012changedetection, wang2014cdnet} and three remote sensing image data sets SVCD \cite{lebedev2018change},LEVIR-CD \cite{chen2020spatial}, and SYSU-CD \cite{shi2021deeply}. 

\textbf{CDnet-2014} consists of 11 scenes, including badWeather, baseline, cameraJitter, dynamicBackground, intermittentObjectMotion, lowFramerate, nightVideos, PTZ, shadow, thermal, and turbulence. Each scene consists of 4-6 videos that are converted into image frames for change detection. Some of these frames are manually annotated with the changed foreground and the unchanged background. In this paper, four long videos are used to evaluate trend prediction, including badminton of cameraJitter, blizzard of badWeather, copyMachine of shadow and office of baseline. For each video, the bi-temporal images are constructed with 15 annotated image frames at intervals, while their change and trend labels can be generated according to three trend definitions in Section \ref{section:Trend_prediction}. Finally, this reconstructed data set are randomly split into training, validation, and testing sets with the ratio of 2:1:1. It is worth noting that trend labels will not be used during training of the proposed weakly-supervised TCD model, which are only used to evaluate model and compare the results visually.

\textbf{SVCD} is obtained from Google Earth, including seven paired real images with the size of $4725 \times 2700$ and four paired synthetic images with the size of $1900 \times 1000$. The synthetic images are generated by manually adding objects to image. All paired images are cropped into 16000 paired images, where 10000/3000/3000 paired images are employed as the training/validation/testing data sets.	

\textbf{LEVIR-CD} is composed of 637 paired remote sensing images with the size of $1024 \times 1024$, which focuses on the building changes. In this paper, we crop one paired images to sixteen sub-images of $256 \times 256$. Based on the division of author to original images, 7120/1024/2048 paired images are obtained as the training/validation/testing data sets.

\textbf{SYSU-CD} is a remote sensing data set obtained from Hong Kong, which contains 20000 bi-temporal images with the size of $256 \times 256$. The training/validation/testing data sets have 10000/4000/4000 paired images, respectively.

\textbf{Evaluate Metrics.} In this paper, five metrics are employed to evaluate model performance, including precision ({P}), recall ({R}), F1-score ({F}), intersection over union ({IoU}), and overall accuracy ({OA}). The above metrics can be defined as, 
\begin{equation}
	\begin{aligned}
		\textbf{P} &= \frac{ \textit{TP}}{ \textit{TP + FP}},\ 
		\textbf{R} = \frac{ \textit{TP}}{ \textit{TP + FN}},\ 
		\textbf{F} = \frac{2\cdot\textbf P\cdot\textbf R}{\textbf P + \textbf R},\\
		\textbf{IoU} &= \frac{\textit {TP}}{\textit {TP + FP + FN}},\ 
		\textbf{OA} = \frac{\textit {TP + TN}}{\textit {TP + FP + TN + FN}},
	\end{aligned}
\end{equation}
where \textit{TP}, \textit{FP}, \textit{TN} and \textit{FN} denote the number of truth positives, false positives, true negatives and false negatives. To evaluate the trend prediction, the \textit{TP}, \textit{FP}, \textit{TN} and \textit{FN} of three change trends would be counted separately.

\subsection{Implementation Details}
\textbf{Hardware and Training Setup.}
The proposed method is implemented using Pytorch on a high-performance computer with two GeForce RTX 3090 GPUs and an Inter Xenon Silver 4214R processor. The model parameters are initialized randomly, and Adam optimizer is used to optimize them during training. The batch size and epoch number are set to 16 and 200. The learning rate is initially set to $10^{-3}$, and then decreased by a factor of 10 every 60 epochs. The input size of the image remains unchanged. To simply highlight feature channel, the temperature of the softmax layer is set to $0.1$. \textbf{Data Augmentation} is applied to improve the model generalization, including random manners of cropping, rotation and flipping. All augmentation strategies are performed in the same way on two paired images.

\subsection{Result Analyses}
Under the supervision of simple change label, the proposed weakly-supervised model can predict both change and trend maps in the GCD and TCD branches simultaneously after inputting bi-temporal images. To assess the effectiveness of model's two branches, their visual and numerical results will be analyzed in turn.

For visual results, we select six examples from CDnet-2014 and SVCD, and show their change and trend prediction maps in Fig. \ref{fig:feature_show}. Particularly, the background channel of $\mathbf{F}^1_\mathbf{I}$ and $\mathbf{F}^2_\mathbf{I}$ are shown to visualize the background information extracted by the proposed strategic approach.
For the GCD task, six predicted change maps are very close to the corresponding change labels, which demonstrates the effectiveness of the model's GCD branch. Moreover, it can be shown that the softmatch distance successfully measures the difference between $\mathbf{F}_{\mathbf{C}}^{1}$ and $\mathbf{F}_{\mathbf{C}}^{2}$, indicating a promising capacity for predicting change map. In terms of the weakly-supervised TCD task, the predicted trend map of CDnet-2014 are consistent with the trend labels. Additionally, despite the lack of trend labels for reference, the SVCD results demonstrate a promising level of accuracy that can be directly observed. Therefore, both the encouraging performance of the model's TCD branch and the softmatch distance's ability to enhance the fore-/back-ground directivity of features are verified simultaneously. By observing Fig. \ref{feature_show_t1} and \ref{feature_show_t2}, it is obvious that the background information of images \textit{T1} and \textit{T2} is extracted almost perfectly. This is another crucial factor contributing to the accurate trend maps that are achieved.

\begin{table}[htbp]
	\centering
	\scriptsize
	\renewcommand\arraystretch{1}
	\setlength{\tabcolsep}{2.2mm}{
		\begin{tabular}{c|c|ccccc}
			\toprule
			\textbf{Video}&\textbf{Type} & \textbf{P} & \textbf{R} & \textbf{F}  & \textbf{IoU} & \textbf{OA} \\
			\midrule
			\multirow{4}[0]{*}{badminton}&C&93.59 &84.52 &88.83&79.90&98.03\\
			&A&81.63&73.00&77.08&62.70&98.69\\
			&D&83.60&61.35&70.77&54.76&98.29\\
			&T&90.77&64.32&75.29&60.37&98.79\\
			\midrule
			\multirow{4}[0]{*}{blizzard}&C&93.29&97.18&95.20&90.83&99.90\\
			&A&96.54&92.86&94.67&89.88&99.73\\
			&D&91.51&94.87&93.16&87.19&99.64\\
			&T&91.67&72.85&81.19&68.33&99.98\\
			\midrule
			\multirow{4}[0]{*}{copyMachine}&C&92.71&94.60&93.65&88.05&98.52\\
			&A&89.96&87.88&88.91&80.03&99.16\\
			&D&89.73&90.06&89.90&81.65&99.21\\
			&T&84.03&88.60&86.25&75.83&98.94\\
			\midrule
			\multirow{4}[0]{*}{office}&C&95.88&97.93&96.89&93.98&99.33\\
			&A&78.97&80.73&79.84&66.44&99.50\\
			&D&82.25&87.50&84.79&73.60&99.56\\
			&T&96.80&94.88&95.83&91.99&99.33\\ 	
			\bottomrule
		\end{tabular}%
	}\vspace{2pt}
	\caption{Evaluation results of trend prediction on four videos of CDnet-2014 data set. Here, ``C'', ``A'', ``D'' and ``T'' indicate ``Change'', ``Appear'', ``Disappear'' and ``Transform'', respectively.}
	\label{table:trend_predicition}%
	\vspace{-10pt}
\end{table}%

The numerical results listed in Table \ref{table:trend_predicition} further illustrate that the trend prediction of four videos are performed successfully. Specifically, the comprehensive metrics {F} and OA values for ``blizzard'', ``copyMachine'' and ``office'' are above $80\%$ and $98\%$, respectively. In summary, the proposed weakly-supervised method, which relies on the simple change labels, effectively conducts the peer prediction of GCD task and the upgrade prediction of the weakly-supervised TCD task, thereby proving its effectiveness.

\subsection{Comparative Studies}
Since we are unable to find existing models for weakly-supervised TCD, the GCD branch results are adopted to conduct comparative studies. If our proposed method surpasses the state-of-the-art approaches in GCD task, it will demonstrate its superiority, particularly the effectiveness of the softmatch distance. Specifically, thirteen comparative methods are employed, including FC-EF \cite{daudt2018fully}, FC-SC \cite{daudt2018fully}, FC-SD \cite{daudt2018fully}, CDNet \cite{alcantarilla2018street}, STANet \cite{chen2020spatial}, BIT \cite{chen2021remote}, CLNet \cite{zheng2021clnet}, DSAMNet \cite{shi2021deeply}, ESCNet \cite{zhang2021escnet}, IFN \cite{zhang2020deeply}, ISNet \cite{cheng2022isnet}, SNUNet \cite{fang2021snunet}, DARNet \cite{li2022densely}. The corresponding experimental results on three data sets SVCD, SYSU, and LEVIR-CD are listed in Table \ref{table:compard_svcd}, \ref{table:compard_sysu}, and \ref{table:compard_levir}, respectively. 

According to the results, the proposed method achieves the highest values for all three comprehensive metrics (F, IoU, and OA) on the three datasets. Notably, the model shows significant improvement than other methods on the SVCD data set. Compared to the second-ranking method DARNet, {F}, {IoU} and {OA} are increased by 1.1\%, 2.09\% and 0.23\%, respectively. In addition, our method outperforms most others for the P and R metrics, despite their typically inverse relationship. For SYSU-CD and LEVIR-CD data sets, the encouraging performance are acquired, which reflects in the balanced P and R and highest {F}, {IoU}, and {OA}. These experiments provide further evidence of the effectiveness of the softmatch distance.

\subsection{Ablation Studies}
As a pluggable module, softmatch distance can be easily replaced with other common distances (Cosine and Euclidean distances) to verify its contribution to our model. We decided to perform this replacement in the GCD branch for ablation studies, since the weakly-supervised TCD task must be supported by using the softmatch distance. Based on this, ablation experiments are conducted on three data sets SVCD, LEVIR-CD and SYSU-CD, whose results are listed in Table \ref{table:ablation}. According to these results, it is easy find that softmatch distance outperforms Euclidean and cosine distances for all five evaluate metrics on the SVCD data set. Although P and R do not reach the best scores simultaneously on both LEVIR-CD and SYSU-CD, the fact that three comprehensive metrics F, IoU and OA achieve the highest values demonstrates the superior performance of softmatch distance. Therefore, the contribution of softmatch distance to our model is verified successfully.

\begin{table}[htbp]
	\centering
	\scriptsize
	\renewcommand\arraystretch{1}
	{
		\begin{tabular}{c|ccccc}
			\toprule
			\textbf{Method} & \textbf{P} & \textbf{R} & \textbf{F} & \textbf{IoU} & \textbf{OA} \\
			\midrule
			FC-EF \cite{daudt2018fully} & 89.87  & 66.49  & 76.43  & 61.85  & 94.74  \\
			\midrule
			CDNet \cite{alcantarilla2018street} & 73.40  & 95.31  & 82.93  & 70.84  & 95.37  \\
			\midrule
			FC-SC \cite{daudt2018fully} & 89.50  & 73.67  & 80.82  & 67.81  & 95.52  \\
			\midrule
			FC-SD \cite{daudt2018fully} & 88.62  & 74.78  & 81.11  & 68.23  & 95.53  \\
			\midrule
			STANet \cite{chen2020spatial} & 87.35  & 95.77  & 91.36  & 84.10  & 97.86  \\
			\midrule
			BIT \cite{chen2021remote}  & 95.31  & 87.31  & 91.13  & 83.71  & 98.00  \\
			\midrule
			CLNet \cite{zheng2021clnet} & 93.30  & 89.80  & 91.52  & 84.36  & 98.03  \\
			\midrule
			DSAMNet \cite{shi2021deeply} & 90.16  & \textbf{98.42}  & 94.10  & 88.87  & 98.41  \\
			\midrule
			ESCNet \cite{zhang2021escnet} & 92.86  & 96.06  & 94.43  & 89.45  & 98.60  \\
			\midrule
			IFN \cite{zhang2020deeply}  & 97.15  & 91.70  & 94.34  & 89.29  & 98.70  \\
			\midrule
			ISNet \cite{cheng2022isnet} & 95.18  & 94.43  & 94.80  & 90.12  & 98.78  \\
			\midrule
			SNUNet \cite{fang2021snunet} & 96.33  & 96.00  & 96.16  & 92.61  & 99.09  \\
			\midrule
			DARNet \cite{li2022densely} & 97.05  & 96.91  & 96.98  & 94.14  & 99.29  \\
			\midrule
			\textbf{Our}   & \textbf{98.18}  & 97.98  & \textbf{98.08}  & \textbf{96.23}  & \textbf{99.52}  \\
			\bottomrule
		\end{tabular}%
	}
	\vspace{2pt}
	\caption{Comparative experimental results on SVCD data set.}
	\vspace{-5pt}
	\label{table:compard_svcd}%
\end{table}%

\begin{table}[htbp]
	\centering
	\scriptsize
	
	\renewcommand\arraystretch{1}
	{
		\begin{tabular}{c|ccccc}
			\toprule
			\textbf{Method} & \textbf{P} & \textbf{R} & \textbf{F}  & \textbf{IoU} & \textbf{OA} \\
			\midrule
			STANet \cite{chen2020spatial} & 70.26  & \textbf{86.18} & 77.41  & 63.15  & 88.14  \\
			\midrule
			SNUNet \cite{fang2021snunet} & 78.26  & 74.42  & 76.30  & 61.68  & 89.10  \\
			\midrule
			CLNet \cite{zheng2021clnet} & 79.62  & 74.97  & 77.22  & 62.90  & 89.57  \\
			\midrule
			DSAMNet \cite{shi2021deeply} & 77.11  & 79.90  & 78.48  & 64.59  & 89.67  \\
			\midrule
			BIT \cite{chen2021remote}   & 79.18  & 77.01  & 78.08  & 64.04  & 89.80  \\
			\midrule
			FC-SD \cite{daudt2018fully} & 79.78  & 74.38  & 76.99  & 62.58  & 89.83  \\
			\midrule
			FC-EF \cite{daudt2018fully} & \textbf{84.77} & 68.63  & 75.85  & 61.09  & 89.84  \\
			\midrule
			FC-SC \cite{daudt2018fully} & 83.03  & 71.60  & 76.90  & 62.46  & 89.90  \\
			\midrule
			ISNet \cite{cheng2022isnet} & 80.27  & 76.41  & 78.29  & 64.44  & 90.01  \\
			\midrule
			IFN \cite{zhang2020deeply}   & 82.39  & 73.57  & 77.73  & 63.57  & 90.06  \\
			\midrule
			\textbf{Our}   & 83.21  & 74.91  & \textbf{78.84} & \textbf{65.07} & \textbf{90.52} \\
			\bottomrule
		\end{tabular}%
	}
	\vspace{2pt}
	\caption{Comparative experimental results on SYSU-CD data set.}
	\vspace{0pt}
	\label{table:compard_sysu}%
\end{table}%

\begin{table}[htbp]
	\centering
	\scriptsize
	\renewcommand\arraystretch{1}
	{
		\begin{tabular}{c|ccccc}
			\toprule
			\textbf{Method} & \textbf{P} & \textbf{R} & \textbf{F}  & \textbf{IoU} & \textbf{OA} \\
			\midrule
			FC-SC \cite{daudt2018fully} & 90.76  & 58.95  & 71.47  & 55.61  & 97.60  \\
			\midrule
			FC-SD \cite{daudt2018fully} & 91.97  & 57.84  & 71.02  & 55.06  & 97.60  \\
			\midrule
			IFN \cite{zhang2020deeply}   & 86.95  & 75.24  & 80.67  & 67.61  & 98.16  \\
			\midrule
			DSAMNet \cite{shi2021deeply} & 80.61  & 88.98  & 84.59  & 73.29  & 98.35  \\
			\midrule
			CLNet \cite{zheng2021clnet} & 90.07  & 85.70  & 87.83  & 78.30  & 98.79  \\
			\midrule
			BIT \cite{chen2021remote}   & 89.24  & 89.37  & 89.31  & 80.68  & 98.92  \\
			\midrule
			SNUNet \cite{fang2021snunet} & 91.80  & 88.53  & 90.14  & 82.04  & 99.01  \\
			\midrule
			ISNet \cite{cheng2022isnet} & \textbf{92.46}  & 88.27  & 90.32  & 82.35  & 99.04  \\
			\midrule
			\textbf{Our}   & 91.87  & \textbf{89.59}  & \textbf{90.72} & \textbf{83.01} & \textbf{99.07} \\
			\bottomrule
		\end{tabular}%
	}
	\vspace{2pt}
	\caption{Comparative experimental results on LEVIR-CD data set.}
	\vspace{0pt}
	\label{table:compard_levir}%
\end{table}%

\begin{table}[t]
	\centering
	\scriptsize
	\renewcommand\arraystretch{1}
	{
		\begin{tabular}{c|cccccc}
			\toprule
			\textbf{Data set} & \textbf{Distance} & \textbf{P} & \textbf{R} & \textbf{F} & \textbf{IoU} & \textbf{OA} \\
			\midrule
			\multirow{3}[0]{*}{SVCD} & Euclidean & 97.35  & 97.12  & 97.23  & 94.61  & 99.31  \\
			&Cosine& 97.44  & 97.43  & 97.43  & 94.99  & 99.36  \\
			&{Softmatch} & \textbf{98.18}  & \textbf{97.98}  & \textbf{98.08}  & \textbf{96.23}  & \textbf{99.52}  \\
			\midrule
			\multirow{3}[0]{*}{LEVIR-CD} & Euclidean & 88.09  & \textbf{91.01}  & 89.53  & 81.04  & 98.92  \\
			&Cosine& 89.96  & 90.28  & 90.12  & 82.02  & 98.99  \\
			&{Softmatch} & \textbf{91.87}  & 89.59  & \textbf{90.72} & \textbf{83.01} & \textbf{99.07} \\
			\midrule
			\multirow{3}[0]{*}{SYSU-CD} & Euclidean & \textbf{83.54}  & 73.28  & 78.07  & 64.03  & 90.29  \\
			&Cosine& 78.51  & \textbf{76.44}  & 77.46  & 63.21  & 89.51  \\
			&{Softmatch} & 83.21  & 74.91  & \textbf{78.84} & \textbf{65.07} & \textbf{90.52} \\
			\bottomrule
		\end{tabular}%
	}
	\vspace{2pt}
	\caption{Ablation experimental results of three distance metrics on three data sets}
	\vspace{0pt}
	\label{table:ablation}%
\end{table}%

\section{Conclusion}
In this paper, we introduce a novel softmatch distance, which is used to construct a weakly-supervised TCD branch in a simple GCD model. The modified model consists of three components: feature extraction, GCD branch and weakly-supervised TCD branch. Here, softmatch distance is used in the GCD branch to measure feature difference and generate the change map, while also enhancing the fore-/back-ground directivity of features in TCD branch to predict the trend map. Moreover, a strategic approach is developed to explore background information from bi-temporal images, which is essential for weakly-supervised TCD task. The promising experimental results demonstrate the effectiveness of our proposed method.

{\small
\bibliographystyle{ieee_fullname}
\bibliography{egbib}
}

\end{document}